\documentclass{ecai} 

\usepackage{latexsym}
\usepackage{amssymb}
\usepackage{amsmath}
\usepackage{amsthm}
\usepackage{booktabs}
\usepackage{enumitem}
\usepackage{graphicx}
\usepackage{color}
\usepackage[ruled,vlined]{algorithm2e}
\usepackage{multirow}
\usepackage{siunitx}
\newcommand{\BibTeX}{B\kern-.05em{\sc i\kern-.025em b}\kern-.08em\TeX}

\newcommand{\stdcell}[2]{#1{\scriptsize~$\pm$~#2}}

\begin{document}


\begin{frontmatter}



\title{Low-Complexity Inference in Continual Learning via Compressed Knowledge Transfer}


\author[A]{\fnms{Zhenrong}~\snm{Liu}\thanks{Corresponding Author. Emails: zhenrong.liu@outlook.com, \{janne.m.huttunen,mikko.honkala\}@nokia-bell-labs.com}}
\author[A]{\fnms{Janne M. J.}~\snm{Huttunen}}
\author[A]{\fnms{Mikko}~\snm{Honkala}} 

\address[A]{Nokia Bell Labs, Espoo, Finland}


\begin{abstract}
    Continual learning (CL) aims to train models that can learn a sequence of tasks without forgetting previously acquired knowledge. 
    A core challenge in CL is balancing stability—preserving performance on old tasks—and plasticity—adapting to new ones. 
    Recently, large pre-trained models have been widely adopted in CL for their ability to support both, offering strong generalization for new tasks and resilience against forgetting. 
    However, their high computational cost at inference time limits their practicality in real-world applications, especially the ones requiring low latency or energy usage.
    To address this issue, we explore model compression techniques—pruning and knowledge distillation (KD)—and propose two efficient frameworks tailored for class-incremental learning (CIL), a challenging CL setting where task identities are unavailable during inference. 
    The pruning-based framework includes pre- and post-pruning strategies that apply compression at different training stages. 
    The KD-based framework adopts a teacher-student architecture, where a large pre-trained teacher transfers downstream-relevant knowledge to a compact student. Extensive experiments on multiple CIL benchmarks demonstrate that the proposed frameworks achieve a better trade-off between accuracy and inference complexity, consistently outperforming strong baselines. 
    We further analyze the trade-offs between the two frameworks in terms of accuracy and efficiency, offering insights into their use across different scenarios.
\end{abstract}

\end{frontmatter}


\section{Introduction}

Continual learning (CL) requires models to acquire knowledge from multiple tasks in a sequential manner~\cite{9349197,10444954}.
However, neural networks are prone to catastrophic forgetting~\cite{goodfellow2015empiricalinvestigationcatastrophicforgetting}, where their performance on previously learned tasks deteriorates significantly after learning new ones. 
To overcome this challenge and achieve CL, a model must balance the ability to retain knowledge from old tasks (stability) while adapting to new tasks (plasticity). 

In practical deep learning applications, it is common to leverage pre-trained models and fine-tune them on the target task's dataset~\cite{6909475}. 
This paradigm has also been gaining popularity in CL approaches. 
For instance, large pre-trained models such as ResNets and Vision Transformers have been widely utilized~\cite{ramasesh2022effect}. 
Since these models are trained on large-scale datasets, they acquire strong generalization abilities giving them a significant advantage in terms of plasticity and reducing overfitting. 
Additionally, models with larger parameter capacity are more resilient to catastrophic forgetting, making large pre-trained models beneficial for improving stability.
As a result, large pre-trained models can enhance performance across various CL algorithms, regardless of the specific method employed.

While large pre-trained models can indeed boost the performance of CL algorithms, they come with the drawback of being computationally expensive during inference. 
For instance, replacing a pre-trained ResNet-50 with a ViT-B/16 in experience replay improves performance by approximately 21\% on average across different datasets~\cite{pmlr-v199-ostapenko22a}, but it also more than doubles the per-sample inference computational cost, from 8.2 giga floating point operations (GFLOPs) to 17.6 GFLOPs.
Moreover, scaling to even larger models results in diminishing returns in performance gains relative to the increased cost.

This motivates us to pursue a better trade-off between inference efficiency and model performance.
Specifically, if CL models can benefit from the general knowledge and plasticity provided by pre-trained models while avoiding their inference cost, this would benefit practical CL applications.

Intuitively, this trade-off is achievable because, although large pre-trained models encode extensive general knowledge, not all of it is necessarily relevant to downstream CL tasks.
By effectively extracting and retaining only the essential knowledge for a given CL scenario while compressing the model, we may balance performance and computational efficiency.
This motivates us to explore methods that selectively distill useful knowledge from pre-trained models, allowing CL models to retain the plasticity from pre-trained models while reducing inference costs.

To balance performance and efficiency, various model compression techniques, such as weight pruning~\cite{molchanov2017pruning} and knowledge distillation (KD)~\cite{hinton2014distilling}, have been widely used to accelerate inference and reduce model size while maintaining accuracy. 
However, their application in CL, which inherently relies on sequential training, remains underexplored.

In this paper, we integrate these two techniques, pruning and KD, into CL, aiming to leverage the benefits of pre-trained models while reducing inference costs. 
We focus on the more challenging class-incremental learning (CIL) setting~\cite{9915459}, where models have no access to task identities during inference.
For pruning, we investigate two strategies for compression during training. 
In the pre-pruning strategy, a pre-trained model, fine-tuned on the initial task to align with the downstream data distribution, is pruned based on the initial task data to obtain a compact model to be subsequently used for the entire CIL training. 
In contrast, the post-pruning strategy retains the full model throughout CIL training and performs pruning after each task to produce a lightweight model specifically for inference. 
For KD, we adopt a parallel teacher-student framework, where the teacher is a larger pre-trained model used only during training, and the student is a compact model used for both training and inference.
Throughout CIL training, KD is employed to transfer downstream-relevant knowledge from the teacher to the student, enhancing the student’s ability to retain information from previous tasks while adapting to new ones.
The main contributions of this work are as follows:
\begin{itemize}
    \item The integration of pruning and KD into CIL is investigated, and two frameworks are proposed to reduce inference cost while maintaining strong performance. 
    \item A comprehensive comparison between the two frameworks is conducted, and their respective strengths in accuracy, efficiency, and architectural flexibility are analyzed. 
    \item Extensive experiments across diverse CIL methods and datasets demonstrate that the proposed frameworks consistently deliver significant improvements, showcasing strong robustness and method-agnostic applicability.
\end{itemize}

\section{Related Work}

A wide range of CL methods have been proposed, which can be broadly categorized into four main approaches.
Regularization-based methods, such as EWC~\cite{doi:10.1073/pnas.1611835114} and LwF~\cite{8107520}, add regularization terms like knowledge distillation or weight penalties to preserve knowledge from previous tasks.
Replay-based methods, such as iCaRL~\cite{Rebuffi_2017_CVPR} and SS-IL~\cite{Ahn_2021_ICCV}, store and replay past samples to retain knowledge and mitigate forgetting.
Optimization-based methods~\cite{Wang_2021_CVPR,10446458} explicitly design and manipulate the optimization process to improve stability-plasticity trade-offs in CL. 
Architecture-based methods~\cite{9578633,veniat:hal-03276781} allocate separate model components for different tasks, isolating learned knowledge to minimize interference.  

In addition to the above CL approaches, recent prompt-based approaches~\cite{9878681,wang2022dualprompt} have gained attention for their strong performance. 
These methods introduce task-specific prompts while keeping the pre-trained model largely frozen, enabling efficient adaptation with minimal updates. 
Their effectiveness stems from leveraging the rich knowledge in large pre-trained models, which enhance generalization over training from scratch~\cite{ijcai2024p924}. 
This success has driven a growing focus on pre-trained models in CL research.

Notably, even before the rise of prompt-based methods, earlier works such as Side-Tuning~\cite{zhang2020side} and DLCFT~\cite{shon2022dlcft} had already explored the benefits of pre-trained models in CL. 
These approaches are often classified as representation-based~\cite{10444954}, as they utilize pre-trained features to improve the representational capacity of continual learners.
However, earlier studies primarily used smaller architectures like ResNet, whereas prompt-based methods have adopted larger models, such as Vision Transformers, achieving greater performance gains.
More recently, studies have shown that pre-trained models can benefit a wide range of CL algorithms beyond prompt-based designs~\cite{Lee_2023_WACV,ramasesh2022effect}, highlighting their general applicability in improving performance.
Despite these advantages, their high computational cost remains a major challenge for efficient inference.  

Model compression techniques, such as pruning and KD, have been widely studied in deep learning to improve inference efficiency. 
Pruning~\cite{8237560,frankle2018lottery} removes redundant parameters to create a more compact model while maintaining performance. 
KD~\cite{hinton2014distilling} enables efficient learning by transferring knowledge from a large teacher model to a smaller student model. 
Despite their success in general deep learning tasks, their application in CL remains underexplored.

With the increasing reliance on pre-trained models in CL, the associated inference cost poses a significant challenge.
In this work, we investigate how pruning and KD can be integrated into CL to reduce the inference overhead of pre-trained models while preserving strong performance.

\section{Methodology}

In this section, we introduce the proposed approaches to integrating model compression into CL.
We first define the CL problem setup, then present two frameworks that apply pruning and KD to reduce inference cost.

\subsection{Problem Formulation}

CL requires a model to sequentially learn from multiple tasks. 
Let $\mathcal{D}_t = \{x_i^t, y_i^t\}_{i=1}^{N_t}$ represent the dataset for task $t$, where $x_i^t$ is an input sample and $y_i^t$ is its corresponding label. 
The total number of samples for task $t$ is denoted as $N_t$, and the entire sequence consists of $T$ tasks. 
When learning a new task, the model has access to little or no training data from previous tasks, which may result in catastrophic forgetting. 
In this work, we focus on \textit{CIL}, where task identities $t$ are not available during inference, meaning the model must classify samples among all previously seen classes without knowing which task they belong to.
This is a more challenging scenario than task-incremental learning, where task identities are known at test time.

To illustrate how model updates are performed in CIL, we present one of the most fundamental approaches: regularization-based methods.
The general form of the loss function for task $t$ is~\cite{NEURIPS2023_249f73e0}:
\begin{equation}
    \label{regularization-based loss}
        \mathcal{L}(\mathbf{\Theta}_t; \mathcal{D}_t)\!=\!\!\!\!\!\!\sum_{(x, y) \in \mathcal{D}_t} \!\!\!\!\!\left[ \mathcal{L}_{\mathrm{cls}}(f_{\mathbf{\Theta}_t}(x), y)
        \!+\!\lambda \mathcal{L}_{\mathrm{reg}}(\mathbf{\Theta}_t; \mathbf{\Theta}_{t-1}, x) \right].
\end{equation}
where \( f_{\mathbf{\Theta}_t} \) denotes the neural network parameterized by \( \mathbf{\Theta}_t \), i.e., the model after learning task \( t \). The classification loss $\mathcal{L}_{\mathrm{cls}}$, typically cross-entropy, optimizes the model for plasticity, while the regularization term $\mathcal{L}_{\mathrm{reg}}$ constrains model updates to retain previous knowledge, improving stability.

To further enhance CIL performance, pre-trained models have been widely adopted to replace training \( f_{\mathbf{\Theta}_t} \) from scratch.
However, this comes with increased inference cost. 
To address this, we explore model compression and propose two frameworks—pruning-based and KD-based—that integrate compression into CIL while maintaining competitive performance.

\subsection{Pruning-Based Framework}

\begin{figure*}[ht]
    \centering
    \includegraphics[width=0.98\textwidth, trim=10 10 10 10, clip]{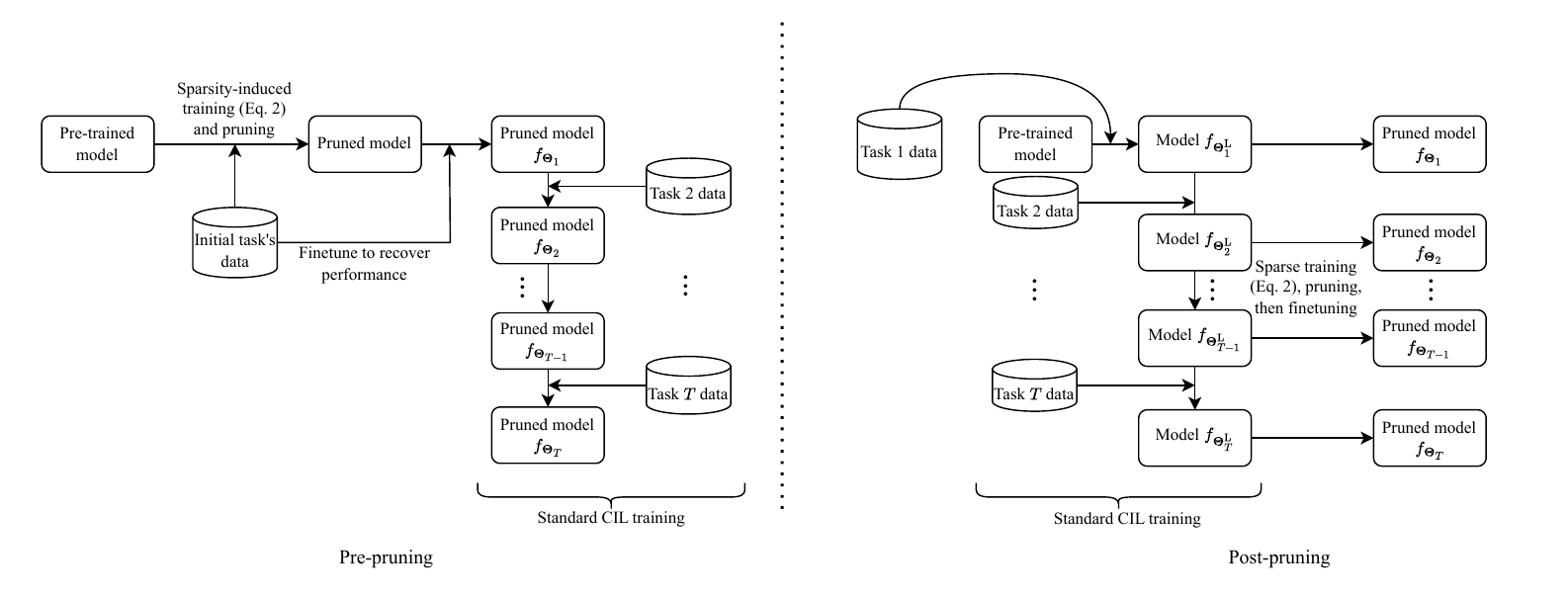}
    \caption{Overview of the proposed pruning-based frameworks for CIL. 
    In pre-pruning, for task $t=1,\ldots,T$, the model $f_{\mathbf{\Theta}_t}$ is initialized from $f_{\mathbf{\Theta}_{t-1}}$ and trained with the current task data. 
    In post-pruning, $f^{\mathrm{L}}_{\mathbf{\Theta}_t}$ denotes the uncompressed model trained on task $t$, initialized from $f^{\mathrm{L}}_{\mathbf{\Theta}_{t-1}}$. 
    After training, it is pruned to produce $f_{\mathbf{\Theta}_t}$ for inference.} 
    \label{pruning-based framework}
\end{figure*}

In the pre-trained–fine-tuned–pruned paradigm, pruning removes less important parameters from pre-trained models while preserving knowledge relevant to downstream tasks~\cite{Liu_Cai_Guo_Chen_2021}.  
However, unlike general deep learning settings where the entire target dataset is available at once, CIL processes data sequentially, making it infeasible to prune based on the full dataset and thus difficult to preserve complete downstream-relevant knowledge.

To address this challenge and effectively integrate pruning into CIL, we propose two strategies.  
The first, referred to as \emph{pre-pruning}, applies pruning before standard CIL sequential training begins.  
Specifically, the pre-trained model is first fine-tuned on the initial task to align with the distribution of the downstream CIL data.  
This step is necessary due to the domain gap between the pre-training data and the downstream tasks—pruning the model directly may otherwise remove critical downstream-relevant knowledge.  
Once aligned, pruning is performed based on importance of parameters estimated from the first task, aiming to reduce model complexity while preserving useful knowledge.  
Since CIL lacks access to future tasks, preserving the full scope of downstream-relevant knowledge is not feasible under this strategy.  
Fortunately, as later tasks typically belong to the same domain and share common features with the initial one, the knowledge preserved from the first task serves as a reasonable approximation of the broader downstream-relevant knowledge.  
After pruning, the model is further fine-tuned to recover potential performance degradation due to parameter removal, and then trained sequentially following standard CIL procedures.  
By removing redundant parameters early, the model becomes more compact and computationally efficient, aiming to balance inference efficiency and knowledge retention.

The second strategy, termed \emph{post-pruning}, delays pruning until after each task.  
In this strategy, the large pre-trained model undergoes standard CIL training without compression.  
After completing each task, pruning is applied to produce a compact model specifically for inference.  
An advantage of this strategy is that it allows the pruned model to benefit from the downstream-relevant knowledge accumulated in the large pre-trained model up to that point, potentially improving knowledge retention compared to pre-pruning.  
However, it introduces greater training complexity, as sequential training must be performed on the large model and pruning must be repeated for each task.  
Notably, the inference cost for each task remains comparable to that of the pre-pruning strategy.  
The two pruning strategies for CIL are illustrated in Figure~\ref{pruning-based framework}.
Since both pre- and post-pruning are performed outside the CIL training pipeline, the proposed framework is easy to integrate on top of existing CIL methods.

\subsubsection{Pruning Method}

We adopt structured pruning over unstructured pruning, as it removes structural components such as filters (in CNNs) or neurons (in MLPs), enabling computational savings without post-processing~\cite{10643325}.
To guide the pruning process during training, we adopt the same approach as in \cite{8237560} and introduce an additional penalty term on batch normalization scale parameters~\cite{pmlr-v37-ioffe15}, leading to the following objective function:  
\begin{equation}
    \mathcal{L}_{\mathrm{prune}}(\mathbf{\Theta}; \mathcal{D}_t) = \sum_{(x, y) \in \mathcal{D}_t} \mathcal{L}_{\mathrm{cls}}(f_{\mathbf{\Theta}}(x), y) + \mu \sum_{\gamma \in \Gamma} |\gamma|,
\end{equation}
where $\mathcal{L}_{\mathrm{cls}}$ ensures effective task learning, while the second term imposes an $\ell_1$ regularization penalty on the batch normalization scale parameters $\Gamma$, promoting sparsity and leading to channel pruning.  
Since batch normalization is typically applied after the linear transformation, its parameters naturally serve as indicators of neuron importance. 
Specifically, scale parameters with larger magnitudes have a greater influence on layer outputs, making them more crucial for feature propagation.
As a result, neurons associated with these significant parameters are retained in the pruned network, while those with the smallest magnitudes—selected globally across the network based on a specified pruning ratio—are removed.
Following pruning, the model is fine-tuned on the current task’s training data to recover potential performance loss.

\subsection{Knowledge Distillation-Based Framework}

In CIL, KD is commonly used between two networks of the same size to transfer knowledge from previous tasks to the current task model, helping to mitigate forgetting.
However, beyond CIL, KD is widely used for transferring knowledge from a large, well-trained model (teacher) to a smaller model (student) to enhance performance.  
In this subsection, we leverage this broader application of KD in the CIL setting by introducing a teacher-student framework, where the pre-trained model acts as the teacher, guiding a more compact yet effective student model. 

\subsubsection{Teacher-Student Framework}  
To better integrate KD with the CIL, we design a parallel teacher-student framework.
Both models are initialized with pre-trained weights, allowing general knowledge to be incorporated from the start.
The teacher is a larger model used only during training, while the student is a compact model used for both training and inference.
Following standard CIL procedures, both models are trained sequentially, enabling effective knowledge transfer from the teacher to the student.
Specifically, this transfer occurs in two stages.

\begin{figure}[ht]
    \centering
    \includegraphics[width=0.48\textwidth, trim=10 10 5 10, clip]{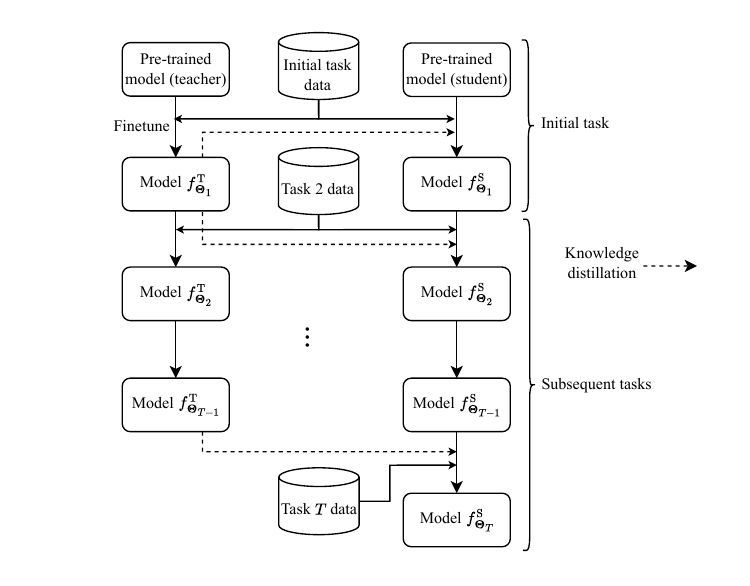}
    \caption{Overview of the proposed KD-based framework for CIL.
    The teacher $f^{\mathrm{T}}_{\mathbf{\Theta}_t}$ is updated via standard CIL training. 
    For task $t=1,\ldots,T$, the student $f^{\mathrm{S}}_{\mathbf{\Theta}_t}$ is initialized from $f^{\mathrm{S}}_{\mathbf{\Theta}_{t-1}}$ and trained on the current task with KD from $f^{\mathrm{T}}_{\mathbf{\Theta}_{t-1}}$ to enhance stability.} 
    \label{KD-based framework}
\end{figure}

In the first task, the teacher is fine-tuned on the task data and then used to train the student via KD, enabling the student to acquire refined, downstream-relevant knowledge.
Since stability is not a concern in the first task of CIL, this training sequence is designed to maximize the student’s plasticity. 
In subsequent tasks, however, forgetting becomes a challenge due to limited or no access to previous task data.
To address this, the student is trained first with KD from the teacher trained on the previous task, helping to mitigate catastrophic forgetting. 
Afterward, the teacher undergoes standard CIL training on the new task.\footnote{We also experimented with the student distilling from the teacher updated on the current task, but it performed slightly worse than using the previous-task teacher. Thus, we adopt and report the latter.}
The proposed KD-based framework for CIL is illustrated in Figure~\ref{KD-based framework}.
In conventional CIL, the student relies solely on its previous-task model to prevent forgetting~\cite{8107520}. 
In contrast, the proposed framework replaces the previous-task student with a progressively updated teacher,  which better prevents the forgetting of downstream-relevant and previously learned task knowledge across sequential training.

\subsubsection{Objective Function}  
\label{sec:KD+LwF}
In this subsection, we illustrate the integration of the proposed KD-based framework on top of LwF—a representative regularization-based CIL method—as an example (see Appendix~\ref{app:kd_icarl_ssil} for a brief description of LwF). 

During the first task, the teacher model is fine-tuned by minimizing the classification loss:
\begin{equation}
    \label{eq:teacher_init}
    \mathcal{L}^\text{T}_{\text{init}} = \sum_{(x, y) \in \mathcal{D}_1}\mathcal{L}_{\mathrm{cls}}(f^{\mathrm{T}}_{\mathbf{\Theta}_1}(x), y) ,
\end{equation}  
where $f^{\mathrm{T}}_{\mathbf{\Theta}}$ denotes the teacher network.
Then, the student is trained using both ground-truth supervision and knowledge distillation, where its logits are aligned with those of the teacher via a KL-divergence-based loss:
\begin{equation}
    \label{eq:student_init}
    \begin{aligned}
        \mathcal{L}^\text{S}_{\text{init}} = &\sum_{(x, y) \in \mathcal{D}_1}\left[\mathcal{L}_{\mathrm{cls}}(f^{\mathrm{S}}_{\mathbf{\Theta}_1}(x), y) \right. \\
        &\left.+\lambda_{\text{init}} \text{KL} \left( \mathbf{p}(f^{\mathrm{T}}_{\mathbf{\Theta}_1}(x), \tau) \parallel \mathbf{p}(f^{\mathrm{S}}_{\mathbf{\Theta}_1}(x), \tau) \right)\right],
    \end{aligned}
\end{equation}  
where $f^{\mathrm{S}}_{\mathbf{\Theta}}$ is the student network, and $\lambda_{\text{init}}$ controls the weight of the KD loss. 
The $i$-th element of soft targets $\mathbf{p}$ are computed using temperature-scaled softmax: 
\begin{equation}
    p^{(i)}(f_{\mathbf{\Theta}}(x), \tau) = \frac{\exp(f_{\mathbf{\Theta}}(x)^i / \tau)}{\sum_j \exp(f_{\mathbf{\Theta}}(x)^j / \tau)},
\end{equation}   
where $\tau$ is a temperature parameter that smooths the output distribution.

For subsequent tasks, the student is trained with a modified loss, where KD is applied only to the logits of previously learned tasks. 
The training loss at this stage is given by:
\begin{equation}
    \label{eq:student_subseq}
    \begin{aligned}
    \mathcal{L}^\text{S}_{\text{sub}}=&\!\!\!\!\sum_{(x, y) \in \mathcal{D}_t}\!\!\!\! \left[\mathcal{L}_{\mathrm{cls}}(f^{\mathrm{S}}_{\mathbf{\Theta}_t}(x), y) \right.\\
    +&\left.\lambda_s \text{KL} \left( \mathbf{p}_{\mathcal{C}_{t-1}}(f^{\mathrm{T}}_{\mathbf{\Theta}_{t-1}}(x), \tau) \parallel \mathbf{p}_{\mathcal{C}_{t-1}}(f^{\mathrm{S}}_{\mathbf{\Theta}_t}(x), \tau) \right) \right],
    \end{aligned}
\end{equation}  
where $\mathcal{C}_{t-1}$ represents the set of classes from previous tasks, $\mathbf{p}_{\mathcal{C}_{t-1}}(\cdot, \tau)$ denotes the logit vector restricted to class set $\mathcal{C}_{t-1}$, and $\lambda_s$ is a scaling factor that balances stability and plasticity.
The second KL term serves as the student’s regularization loss $\mathcal{L}_{\mathrm{reg}}$, as in Equation~(\ref{regularization-based loss}) and consistent with the LwF.
After the student is updated, the teacher is fine-tuned on the current task using the base CIL method (in this case, LwF). The overall training procedure is outlined in Algorithm~\ref{alg:kd_cil}.

While LwF is used here for illustration, the proposed KD-based framework can be readily applied to other CIL methods.
In such cases, the overall pipeline remains unchanged: the teacher is trained according to the selected CIL method, while the student either incorporates an additional KD loss from the teacher or replaces the original KD loss (designed to mitigate forgetting in the base method) with a teacher-driven one.
The integration details for other CIL methods such as iCaRL and SS-IL are described in Appendix~\ref{app:kd_icarl_ssil}. 

\begin{algorithm}[ht]
    \caption{KD-Based CIL Framework}
    \label{alg:kd_cil}
    \KwIn{Pre-trained teacher $f^{\mathrm{T}}_{\mathbf{\Theta}}$ and student $f^{\mathrm{S}}_{\mathbf{\Theta}}$, dataset $\mathcal{D}$ with tasks $\{\mathcal{D}_1, \mathcal{D}_2, ..., \mathcal{D}_T\}$}
    \KwOut{Trained student model $f^{\mathrm{S}}_{\mathbf{\Theta}_T}$}
    
    \textbf{Initial Task ($t=1$)}: \\
    \quad Train teacher on $\mathcal{D}_1$ with classification loss (Eq.~\ref{eq:teacher_init}).\\
    \quad Train student with KD from teacher (Eq.~\ref{eq:student_init}).\\
    
    \For{$t = 2$ \textbf{to} $T$}{
        Train student with KD from previous-task teacher (Eq.~\ref{eq:student_subseq}).\\
        Update teacher using standard CIL training.\\
        }
    
    \Return $f^{\mathrm{S}}_{\mathbf{\Theta}_T}$
\end{algorithm}

\section{Experiments}

\subsection{Datasets}  

\begin{table*}[ht]
    \caption{Performance comparison (ACC and BWT) of baseline methods and KD-enhanced variants across CIFAR-100, Aircrafts, and Cars datasets.}
    \label{tab:combined_results}
    \centering
    \renewcommand{\arraystretch}{1.15}
    \begin{tabular}{l c c c c c c}
        \toprule
        \textbf{Method} 
        & \multicolumn{2}{c}{\textbf{CIFAR-100}} 
        & \multicolumn{2}{c}{\textbf{Aircrafts}} 
        & \multicolumn{2}{c}{\textbf{Cars}} \\
        & ACC & BWT & ACC & BWT & ACC & BWT \\
        \midrule

        LwF~\cite{8107520} 
        & \stdcell{11.81}{1.44} & \stdcell{-43.60}{2.50} 
        & \stdcell{11.62}{1.44} & \stdcell{-27.09}{1.72} 
        & \stdcell{~~8.69}{0.73} & \stdcell{-19.89}{1.18} \\

        Fetril~\cite{petit2023fetril}
        & \stdcell{12.29}{2.50} & \stdcell{~~-9.38}{1.76}
        & \stdcell{13.36}{1.56} & \stdcell{~~-9.80}{0.81}
        & \stdcell{~~9.33}{3.87} & \stdcell{~~-6.75}{3.13} \\

        PASS~\cite{9578909}
        & \stdcell{14.52}{1.50} & \stdcell{~~-5.98}{5.41}
        & \stdcell{12.75}{0.95} & \stdcell{~~-7.50}{0.49}
        & \stdcell{~~8.25}{1.20} & \stdcell{~~-8.58}{1.67} \\

        FSA~\cite{10378197} 
        & \stdcell{18.43}{1.25} & \stdcell{~~-9.13}{0.39}
        & \stdcell{\textbf{33.57}}{0.66} & \stdcell{-10.80}{0.79}
        & \stdcell{\textbf{30.68}}{0.45} & \stdcell{-27.90}{1.38} \\

        iCaRL (MLP)~\cite{Rebuffi_2017_CVPR} 
        & \stdcell{19.49}{0.62} & \stdcell{-52.83}{0.57}
        & \stdcell{17.30}{2.76} & \stdcell{-48.45}{3.60}
        & \stdcell{17.31}{3.18} & \stdcell{-52.81}{2.15} \\

        iCaRL (NCM)~\cite{Rebuffi_2017_CVPR} 
        & \stdcell{27.75}{0.57} & \stdcell{-36.25}{1.43}
        & \stdcell{21.98}{2.64} & \stdcell{-38.44}{2.82}
        & \stdcell{23.81}{4.03} & \stdcell{-36.03}{1.28} \\

        SS-IL (MLP)~\cite{Ahn_2021_ICCV} 
        & \stdcell{23.32}{2.33} & \stdcell{-34.23}{2.25}
        & \stdcell{17.05}{2.12} & \stdcell{-48.49}{3.33}
        & \stdcell{18.12}{2.47} & \stdcell{-28.90}{2.32} \\

        SS-IL (NCM)~\cite{Ahn_2021_ICCV} 
        & \stdcell{27.67}{2.33} & \stdcell{-19.77}{2.25}
        & \stdcell{22.32}{2.25} & \stdcell{-38.07}{3.00}
        & \stdcell{18.16}{2.77} & \stdcell{-23.35}{2.08} \\

        WA~\cite{9156766}
        & \stdcell{34.23}{0.23} & \stdcell{-34.64}{0.76}
        & \stdcell{22.82}{3.03} & \stdcell{-28.90}{1.09}
        & \stdcell{16.38}{2.89} & \stdcell{-29.42}{1.77} \\

        Foster~\cite{wang2022foster}
        & \stdcell{34.86}{0.75} & \stdcell{-36.93}{0.59}
        & \stdcell{27.58}{8.48} & \stdcell{-20.00}{3.23}
        & \stdcell{10.91}{1.62} & \stdcell{~~-2.47}{0.60} \\

        LWN~\cite{Tao2024}
        & \stdcell{\textbf{35.18}}{2.61} & \stdcell{~~-3.83}{1.85}
        & \stdcell{~~8.66}{0.94} & \stdcell{~~-1.01}{0.69}
        & \stdcell{~~6.18}{0.58} & \stdcell{~~-1.33}{0.60} \\
        
        \midrule
        Proposed + LwF / LWN
        & \stdcell{40.19}{1.50} & \stdcell{~~-0.14}{2.57}
        & \stdcell{40.45}{1.79} & \stdcell{-17.60}{2.54}
        & \stdcell{48.04}{1.43} & \stdcell{-27.10}{1.70} \\

        Proposed + iCaRL (MLP) 
        & \stdcell{38.94}{0.84} & \stdcell{-45.62}{1.54}
        & \stdcell{38.59}{1.04} & \stdcell{-55.00}{2.29}
        & \stdcell{38.84}{1.80} & \stdcell{-58.79}{2.45} \\

        Proposed + iCaRL (NCM) 
        & \stdcell{46.75}{0.75} & \stdcell{-13.66}{0.69}
        & \stdcell{\textbf{53.56}}{0.78} & \stdcell{-24.16}{1.68}
        & \stdcell{54.06}{1.12} & \stdcell{-20.34}{2.59} \\

        Proposed + SS-IL (MLP) 
        & \stdcell{\textbf{49.77}}{1.10} & \stdcell{~~~6.27}{2.61}
        & \stdcell{46.47}{1.83} & \stdcell{~~~9.31}{4.96}
        & \stdcell{40.84}{1.62} & \stdcell{~~-8.37}{7.67} \\

        Proposed + SS-IL (NCM) 
        & \stdcell{47.76}{0.71} & \stdcell{-10.67}{1.07}
        & \stdcell{52.90}{1.87} & \stdcell{-18.76}{1.62}
        & \stdcell{\textbf{54.22}}{1.20} & \stdcell{-16.48}{2.52} \\
        \bottomrule
    \end{tabular}
\end{table*}

We use pre-trained weights obtained from a large dataset in a related domain—ImageNet, in our case.
To ensure that the CIL task data is unseen during pre-training, we avoid using ImageNet-based datasets.
Our evaluation is conducted on three benchmark datasets: CIFAR-100, FGVC Aircraft, and Cars.
Following the categorization in \cite{pmlr-v199-ostapenko22a}, we classify datasets as in-distribution (ID) or out-of-distribution (OoD) based on their semantic similarity to ImageNet:
\begin{itemize}
    \item CIFAR-100: Categorized as ID since many of its classes have direct counterparts in ImageNet.  
    \item FGVC Aircraft, Cars: Considered OoD as they contain fine-grained subclasses (e.g., specific car and aircraft models) that do not appear in ImageNet, which only includes broader categories.  
\end{itemize}
To construct the CIL setup, each dataset is divided into 10 sequential tasks such that all task have individual classes. 
If the total number of classes is not divisible by 10, the last task contains fewer classes.

\subsection{Implementation Details.}

The following implementation details apply across all experiments. 
For pre-training, models on CIFAR-100 were initialized using weights pre-trained on ImageNet-32 \cite{chrabaszcz2017downsampledvariantimagenetalternative} due to its lower image resolution, while models on other datasets used standard ImageNet weights from PyTorch.
All models were trained using SGD with a batch size of 128. 
For replay-based methods, the buffer size was set to 2000 for CIFAR-100, following~\cite{Rebuffi_2017_CVPR,Ahn_2021_ICCV}, and scaled proportionally for other datasets based on the relative size of their training sets.
In the KD-based framework, the KD loss was applied with temperature $\tau = 2$, and scaled by $\lambda_{\text{init}} = 1$ for the first task and $\lambda_s = 10$ for subsequent tasks.
In the pruning-based framework, the $\ell_1$ penalty was weighted by $\mu = 0.1$.
Results are averaged over five runs with different random seeds.
All implementations are based on publicly available code (e.g., PyCIL~\cite{zhou2023pycil}) or the original repositories. Hyperparameters follow the original settings when available, or are fine-tuned on a small validation set otherwise.

\subsection{Metrics}  
We evaluate the proposed frameworks using three key metrics: average accuracy (ACC), backward transfer (BWT), and inference cost.  
ACC measures the final classification performance across all tasks:  
\begin{equation}
    \text{ACC} = \frac{1}{T} \sum_{t=1}^{T} a_{T,t},
\end{equation}
where $T$ is the total number of tasks, and $a_{T,t}$ is the accuracy on task $t$ after training on all $T$ tasks.  
ACC serves as the primary indicator of overall performance.
BWT quantifies the effect of learning new tasks on previous ones:  
\begin{equation}
    \text{BWT} = \frac{1}{T-1} \sum_{t=1}^{T-1} (a_{T,t} - a_{t,t}),
\end{equation}
where $a_{t,t}$ is the accuracy on task $t$ immediately after training, and $a_{T,t}$ is its accuracy after all tasks are learned. 
A positive BWT indicates beneficial knowledge transfer, while a negative value reflects forgetting.  
When ACC is comparable, a higher BWT is preferred.
Inference cost measures the computational efficiency of a model based on the FLOPs required for a single forward pass. 
A lower FLOPs count indicates higher model efficiency.  

\subsection{Experimental Results}

\subsubsection{KD-Based Framework}
\label{sec:kd-based experiment}

Table~\ref{tab:combined_results} presents the performance across multiple datasets for several representative CIL methods. 
The proposed KD framework is applied to LwF, iCaRL, and SS-IL. 
Integration with LwF is detailed in Section~\ref{sec:KD+LwF}, while adaptations for iCaRL and SS-IL are provided in Appendix~\ref{app:kd_icarl_ssil}. 
Notably, the proposed KD-augmented versions of LwF and LWN are equivalent, as LWN can be regarded as an extension of LwF that incorporates a larger (non-pre-trained) teacher model to mitigate forgetting.
All methods use MobileNetV2 as the backbone and adopt a single-layer MLP as the default classifier. 
For iCaRL and SS-IL, additional results using a nearest class mean (NCM) classifier~\cite{6517188} are also reported. 
In the proposed KD framework, the teacher is a ResNet-34 and the student adopts MobileNetV2. 
All methods share the same inference model, MobileNetV2, which requires only 0.013 GFLOPs per inference on CIFAR-100.

On the ID dataset (CIFAR-100), integrating the proposed KD framework into various CIL methods yields substantial ACC improvements—typically around 20 percentage points—over their respective baselines.
In terms of BWT, all methods benefit from KD, showing consistent improvements over their original versions.
Among the baselines, LWN achieves the highest ACC due to the enhanced forgetting resistance provided by its larger teacher, which is also reflected in its relatively low BWT. 
Building upon this strong baseline, the proposed KD framework further improves ACC by approximately 5 percentage points.
These results confirm that the downstream-relevant knowledge distilled from the teacher contributes to both plasticity (higher ACC) and stability (improved BWT), highlighting the effectiveness of the proposed framework.

On the OoD datasets (Aircrafts and Cars), the KD framework also consistently outperforms all baselines in terms of ACC. 
This suggests that even under semantic shift, the teacher is able to transfer useful downstream-relevant knowledge to the student.
It is worth noting that compared to other baselines, LWN performs poorly on the OoD datasets. 
This can be attributed to data scarcity: while CIFAR-100 offers 500 training images per class, FGVC Aircraft and Stanford Cars contain only around 64 and 41 images per class, respectively. 
The limited data makes it difficult for LWN to train its large teacher from scratch effectively. 
This is reflected in its low ACC and small absolute BWT: while the latter suggests low forgetting, it does not translate into high ACC.
This indicates that the method fails to achieve high task-wise accuracy after learning each task.
Although ACC is the primary metric, KD-based methods may show lower BWT on OoD datasets, not due to increased forgetting, but because they achieve higher task-wise accuracy after each task, leading to larger drops at the end of CIL.
In contrast, baseline methods usually reach lower task-wise accuracy, resulting in smaller drops and thus higher BWT scores.
The above analysis is supported by the figures showing ACC progression over incremental tasks on OoD datasets, provided in Appendix~\ref{app:ood_acc} with corresponding explanations.

\begin{figure}[ht]
    \centering
    \includegraphics[width=0.483\textwidth, trim=7 6 7 6, clip]{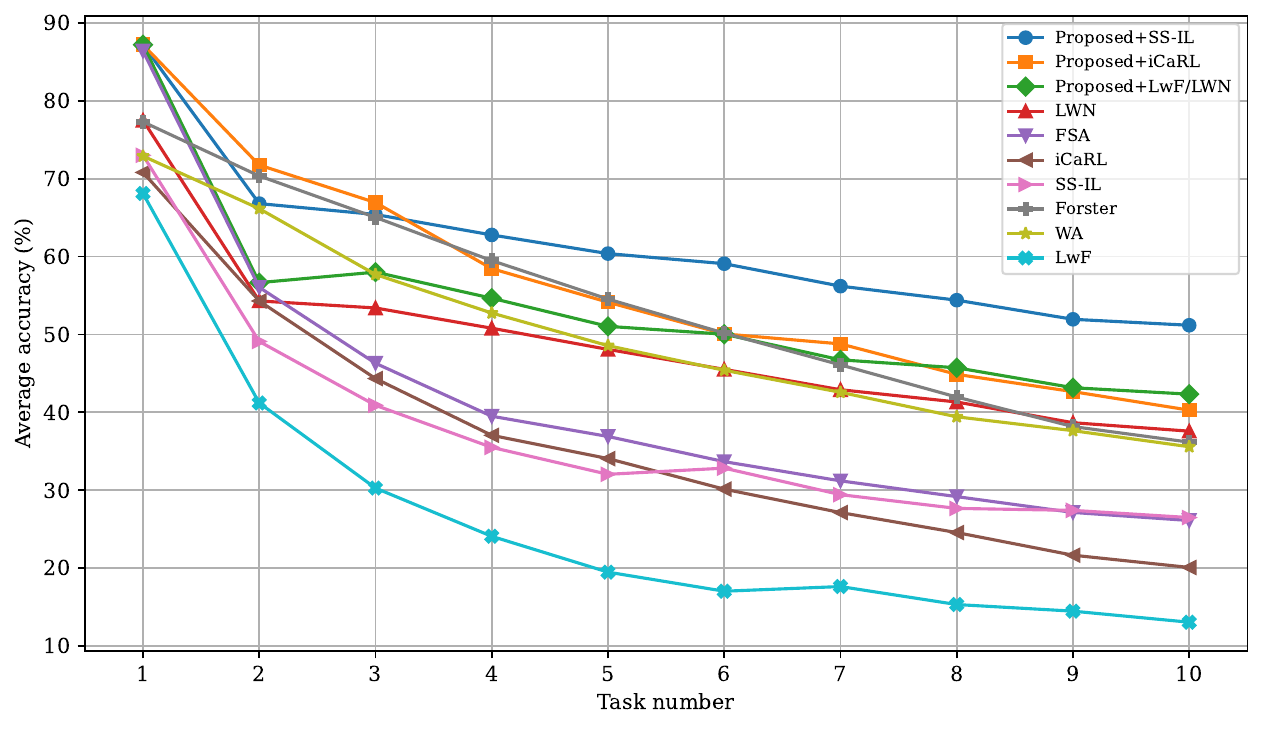}
    \caption{ACC vs. task number on CIFAR-100 for the proposed KD framework.} 
    \label{task_accuracy_cifar100-10}
\end{figure}

Figure~\ref{task_accuracy_cifar100-10} illustrates the ACC progression on CIFAR-100 over incremental tasks, comparing the baseline methods and their KD-based counterparts using the MLP classifier (shown for one random seed). 
After the first task, all KD-enhanced methods already surpass their respective baselines, demonstrating the immediate benefit of teacher-guided learning. 
Throughout the sequence of tasks, the KD-based variants consistently maintain higher ACC compared to their non-KD versions and the strongest baseline, further confirming the robustness and reliability of the proposed KD framework.
The ACC progression curves for OoD datasets are provided in Appendix~\ref{app:ood_acc}, where similar trends can be observed.

\subsubsection{Pruning-Based Framework}

\begin{table}[ht]
    \caption{Comparison of pre- and post-pruning strategies with various pruning ratios on CIFAR-100.}
    \centering
    \begin{tabular}{l@{\hspace{3pt}}c@{\hspace{5pt}}c@{\hspace{5pt}}S[table-format=2.2]@{\hspace{5pt}}c}
    \toprule
    \textbf{Strategy} & \textbf{Pruning Ratio} & \textbf{ACC} & \textbf{Parameters (M)} & \textbf{FLOPs (G)} \\
    \midrule
    Pre-trained & --- & 47.33 & 21.28 & 2.32 \\
    \midrule
    \multirow{5}{*}{Pre-pruning} 
        & 30\% & 42.25 & 13.95 & 1.26 \\
        & 40\% & 44.33 & 11.31 & 1.07 \\
        & 50\% & 42.50 & 8.38  & 0.90 \\
        & 60\% & 40.98 & 5.87  & 0.76 \\
        & 70\% & 37.37 & 3.89  & 0.58 \\
    \midrule
    \multirow{4}{*}{Post-pruning} 
        & 30\% & 30.19 & 15.48 & 1.43 \\
        & 40\% & 20.36 & 12.37 & 1.14 \\
        & 50\% & 23.51 & 10.22 & 0.93 \\
        & 60\% & 17.48 & 7.30  & 0.77 \\
    \bottomrule
    \end{tabular}
    \label{tab:pruning_results}
\end{table}

Table~\ref{tab:pruning_results} presents the performance of the proposed pruning strategies under various pruning ratios on the CIFAR-100 dataset.
We use LwF as the base method to evaluate how different pruning strategies perform under varying pruning ratios.
All pruning experiments are conducted on the pre-trained ResNet-34.
To assess the trade-off between performance and efficiency, we report ACC, parameter count, and single-sample FLOPs.
It can be observed that pruning significantly reduces both the number of parameters and the inference cost.
For example, under a 40\% pruning ratio with the pre-pruning strategy, ACC drops by only 3 percentage points, while the parameter count and FLOPs are both reduced by approximately half.
This demonstrates that pruning offers an effective trade-off between model performance and efficiency.

\begin{figure}[ht]
    \centering
    \includegraphics[width=0.484\textwidth, trim=7 7 7 7, clip]{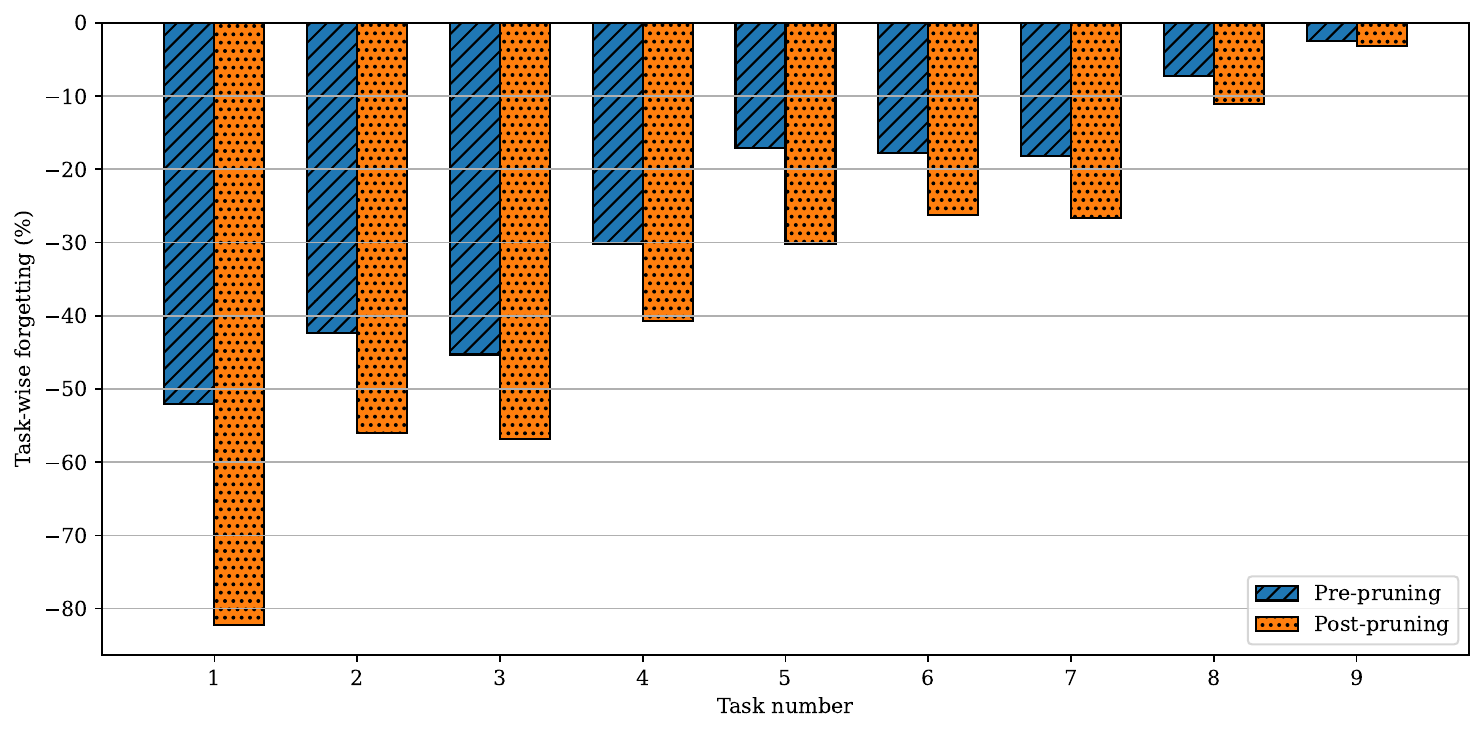}
    \caption{Comparison of task-wise forgetting between pre-pruning and post-pruning. Here, task-wise forgetting is measured as $a_{T,t} - a_{t,t}$, and BWT represents the average over all tasks.}
    \label{fig:forgetting_post_pre_pruning}
\end{figure}

When comparing the two proposed strategies, pre-pruning consistently outperforms post-pruning in terms of performance metrics.
Given its lower training complexity, pre-pruning is the more favorable strategy for integrating pruning into CIL.
The superior performance of pre-pruning in terms of ACC may stem from how pruning interacts with forgetting. 
In the typical training–pruning–fine-tuning paradigm, pruning often leads to some accuracy degradation, but this can usually be recovered through fine-tuning on the original training data~\cite{frankle2018lottery}. 
However, in the post-pruning setting within CIL, pruning-induced degradation impacts both the current and previous tasks. 
Since past data is unavailable or limited—especially in our LwF-based setting without replay—only current-task performance can be recovered through fine-tuning, while earlier forgetting cannot be reversed.
Figure~\ref{fig:forgetting_post_pre_pruning} illustrates this effect, showing that post-pruning causes significantly more forgetting across all tasks compared to pre-pruning.
As a result, post-pruning is more susceptible to unrecoverable loss, leading to worse overall ACC.

Another observation is that increasing the pruning ratio does not lead to a monotonic decline in ACC, as seen in both pre-pruning and post-pruning results.
This observation suggests the existence of a sweet spot where pruning significantly reduces model complexity (e.g., parameters and FLOPs) with minimal ACC degradation, echoing similar patterns observed in the general network pruning literature~\cite{frankle2018lottery,10643325}.

\subsubsection{Comparison of KD and Pruning-Based Framework}

\begin{table}[ht]
    \caption{Comparison of KD- and pruning-based frameworks (with various pruning ratios) on CIFAR-100.}
    \label{tab:KD_pruning_results}
    \centering
    \begin{tabular}{lcS[table-format=2.2] S[table-format=1.3]}
    \toprule
    \textbf{Pruning Ratio} & \textbf{ACC} & \textbf{Parameters (M)} & \textbf{FLOPs (G)} \\
    \midrule
    \textbf{Pre-trained} ResNet34 & 47.33 & 21.28 & 2.32 \\
    \textbf{Pre-trained} MNetV2 & 18.67 & 2.30 & 0.013 \\
    \midrule
    40\% & 44.33 & 11.31 & 1.07 \\
    \textbf{KD} MNetV3 (2×) & 43.96 & 11.83 & 0.046 \\
    \textbf{KD} MNetV3 (1.5×) & 43.77 & 6.76 & 0.027 \\  
    \textbf{KD} MNetV2 (2×) & 42.97 & 8.93 & 0.047 \\
    \textbf{KD} MNetV2 (1.5×) & 42.81 & 5.11 & 0.028 \\
    50\% & 42.50 & 8.38 & 0.90 \\
    30\% & 42.25 & 13.95 & 1.26 \\
    \textbf{KD} MNetV3 & 41.01 & 3.05 & 0.012 \\
    60\% & 40.98 & 5.87 & 0.76 \\
    \textbf{KD} MNetV2 & 40.19 & 2.30 & 0.013 \\
    70\% & 37.37 & 3.89 & 0.58 \\
    \bottomrule
    \end{tabular}
\end{table}

Table~\ref{tab:KD_pruning_results} compares the performance of the proposed pruning-based and KD-based frameworks for the CIFAR-100 dataset. The base CIL method is chosen to be LwF and, for pruning, we focus on the pre-pruning strategy, which consistently outperforms post-pruning in earlier evaluations. 
Pruning is applied to a pre-trained ResNet-34; as in earlier experiments, it also serves as the initial teacher in the KD framework.
For the KD-based framework, we evaluate the default MobileNetV2 student as well as MobileNetV3 and their widened variants (1.5× and 2.0×), with channel widths scaled accordingly.

Overall, the KD-based framework achieves ACC comparable to the pruning-based framework across different student models, with pruning at a 40\% ratio slightly outperforming all KD variants. 
The KD framework, however, delivers significantly larger reductions in FLOPs and/or parameter count while maintaining similar ACC. 
This benefit arises from KD’s architectural flexibility: the teacher and student can adopt entirely different network designs. While pruning must preserve the structure of the original model (e.g., ResNet-34 with standard convolutions), KD allows the use of compact architectures such as MobileNets, which employ depthwise separable convolutions for improved efficiency. 
This makes KD particularly effective in balancing performance with inference cost. 
For example, the KD-based MobileNetV3 (2×) achieves ACC within 4 percentage points of the uncompressed model, while reducing parameter count by half and FLOPs by over 50×.

It is also worth noting that this flexibility comes with additional training complexity. 
KD requires training both the teacher and student models to align them throughout CIL.
In contrast, pruning compresses the model only once before CIL begins, allowing the pruned model to be used throughout, making it a simpler alternative in terms of implementation and training cost.

\subsubsection{Ablation Study}

\begin{table}[ht]
    \caption{Ablation study of the proposed KD-based framework on CIFAR-100.}
    \label{tab:ablation_results}
    \centering
    \begin{tabular}{l l c c }    
    \toprule
    \textbf{Dataset} & \textbf{Method} & \textbf{ACC} & \textbf{BWT} \\
    \midrule
    \multirow{4}{*}{Aircrafts}      
        & LwF                       & \stdcell{11.62}{1.44} & \stdcell{-27.09}{1.72} \\
        & + Pre-trained Only       & \stdcell{35.39}{0.93} & \stdcell{-35.70}{1.82} \\ 
        & + KD Only (LWN)~\cite{Tao2024} & \stdcell{~~8.66}{0.94}  & \stdcell{~~-1.01}{0.69} \\
        & + Proposed (Combined)    & \stdcell{40.45}{1.79} & \stdcell{-17.60}{2.54} \\  
    \midrule
    \multirow{4}{*}{Cars196}        
        & LwF                       & \stdcell{~~8.69}{0.73}  & \stdcell{-19.89}{1.18} \\
        & + Pre-trained Only       & \stdcell{40.88}{0.93} & \stdcell{-33.56}{0.53} \\ 
        & + KD Only (LWN)~\cite{Tao2024} & \stdcell{~~6.18}{0.58}  & \stdcell{~~-1.33}{0.60} \\    
        & + Proposed (Combined)    & \stdcell{48.04}{1.43} & \stdcell{-27.10}{1.70} \\  
    \midrule
    \multirow{4}{*}{CIFAR100}       
        & LwF                       & \stdcell{11.81}{1.44} & \stdcell{-43.60}{2.50} \\
        & + Pre-trained Only       & \stdcell{18.67}{1.52} & \stdcell{-45.76}{2.10} \\
        & + KD Only (LWN)~\cite{Tao2024} & \stdcell{35.18}{2.61} & \stdcell{~~-3.83}{1.85} \\ 
        & + Proposed (Combined)    & \stdcell{40.19}{1.50} & \stdcell{~~-0.14}{2.57} \\   
    \bottomrule
    \end{tabular}
\end{table}

The proposed KD-based framework includes two key components: (1) applying KD from a large teacher to a small student, and (2) incorporating pre-trained knowledge.
These components can be independently integrated into a CIL method—here, we use LwF as the base—to evaluate their individual contributions to performance. 
In Table~\ref{tab:ablation_results}, \emph{KD Only} refers to training both the teacher and student from scratch~\cite{Tao2024}, while \emph{Pre-trained Only} uses a single student initialized with pre-trained weights to incorporate general knowledge.

On CIFAR-100, Pre-trained Only provides a modest improvement over the LwF baseline, whereas KD Only results in a substantially larger gain. 
This is likely because the compact student retains only limited general knowledge due to its small capacity, making it less capable of learning high-quality downstream-relevant features independently.
In contrast, the larger teacher, though trained from scratch, can more effectively learn task-relevant knowledge and transfer it to the student via KD.

On the Cars and Aircraft, however, the trend reverses: KD Only performs worse than LwF, while Pre-trained Only delivers notable gains.
This is due to the limited training data available in these datasets.
Under such low-data regimes, the large teacher struggles to learn effectively, making KD less beneficial or even harmful.
In contrast, a pre-trained student remains more effective despite the scarcity of training data.

Despite these dataset-specific trends, the proposed KD framework consistently achieves the highest ACC across all cases.
This highlights the complementary strengths of the two components: pre-trained knowledge provides a strong general foundation for downstream refinement, while the larger teacher enables high-quality, task-specific learning and effective knowledge transfer through KD.

\section{Conclusion}

In this work, we propose two efficient frameworks for CL by integrating pruning and KD into the CIL setting.
Unlike prior studies that leverage large pre-trained models without considering inference cost, the proposed frameworks aim to retain strong performance by benefiting from pre-trained knowledge while significantly reducing it.
Extensive experiments show that the proposed KD-based framework consistently outperforms standard baselines with the same inference model, achieving a favorable balance between accuracy and efficiency.
The proposed pruning-based framework attains slightly higher accuracy under low pruning ratios, though with less reduction in FLOPs and parameter count compared to KD.
Overall, pruning offers simplicity, while KD supports different architectures for the pre-trained and inference-time models, enabling stronger compression.




\bibliography{references}

\begin{thebibliography}{38}
\providecommand{\natexlab}[1]{#1}
\providecommand{\url}[1]{\texttt{#1}}
\expandafter\ifx\csname urlstyle\endcsname\relax
  \providecommand{\doi}[1]{doi: #1}\else
  \providecommand{\doi}{doi: \begingroup \urlstyle{rm}\Url}\fi

\bibitem[Ahn et~al.(2021)Ahn, Kwak, Lim, Bang, Kim, and Moon]{Ahn_2021_ICCV}
H.~Ahn, J.~Kwak, S.~Lim, H.~Bang, H.~Kim, and T.~Moon.
\newblock {SS-IL}: Separated softmax for incremental learning.
\newblock In \emph{Proceedings of the IEEE/CVF International Conference on Computer Vision (ICCV)}, pages 844--853, 2021.

\bibitem[Cheng et~al.(2024)Cheng, Zhang, and Shi]{10643325}
H.~Cheng, M.~Zhang, and J.~Q. Shi.
\newblock A survey on deep neural network pruning: Taxonomy, comparison, analysis, and recommendations.
\newblock \emph{IEEE Transactions on Pattern Analysis and Machine Intelligence}, 46\penalty0 (12):\penalty0 10558--10578, 2024.

\bibitem[Chrabaszcz et~al.(2017)Chrabaszcz, Loshchilov, and Hutter]{chrabaszcz2017downsampledvariantimagenetalternative}
P.~Chrabaszcz, I.~Loshchilov, and F.~Hutter.
\newblock A downsampled variant of {ImageNet} as an alternative to the {CIFAR} datasets.
\newblock Preprint arXiv:cs/1707.08819, 2017.

\bibitem[De~Lange et~al.(2022)De~Lange, Aljundi, Masana, Parisot, Jia, Leonardis, Slabaugh, and Tuytelaars]{9349197}
M.~De~Lange, R.~Aljundi, M.~Masana, S.~Parisot, X.~Jia, A.~Leonardis, G.~Slabaugh, and T.~Tuytelaars.
\newblock A continual learning survey: Defying forgetting in classification tasks.
\newblock \emph{IEEE Transactions on Pattern Analysis and Machine Intelligence}, 44\penalty0 (7):\penalty0 3366--3385, 2022.

\bibitem[Frankle and Carbin(2018)]{frankle2018lottery}
J.~Frankle and M.~Carbin.
\newblock The lottery ticket hypothesis: Finding sparse, trainable neural networks.
\newblock In \emph{Proceedings of the International Conference on Learning Representations (ICLR)}, 2018.

\bibitem[Girshick et~al.(2014)Girshick, Donahue, Darrell, and Malik]{6909475}
R.~Girshick, J.~Donahue, T.~Darrell, and J.~Malik.
\newblock Rich feature hierarchies for accurate object detection and semantic segmentation.
\newblock In \emph{Proceedings of the IEEE/CVF Conference on Computer Vision and Pattern Recognition (CVPR)}, pages 580--587, 2014.

\bibitem[Goodfellow et~al.(2015)Goodfellow, Mirza, Xiao, Courville, and Bengio]{goodfellow2015empiricalinvestigationcatastrophicforgetting}
I.~J. Goodfellow, M.~Mirza, D.~Xiao, A.~Courville, and Y.~Bengio.
\newblock An empirical investigation of catastrophic forgetting in gradient-based neural networks.
\newblock Preprint arXiv:stat/1312.6211, 2015.

\bibitem[Hinton et~al.(2015)Hinton, Vinyals, and Dean]{hinton2014distilling}
G.~Hinton, O.~Vinyals, and J.~Dean.
\newblock Distilling the knowledge in a neural network.
\newblock Preprint arXiv:stat/1503.02531, 2015.

\bibitem[Ioffe and Szegedy(2015)]{pmlr-v37-ioffe15}
S.~Ioffe and C.~Szegedy.
\newblock Batch normalization: Accelerating deep network training by reducing internal covariate shift.
\newblock In \emph{Proceedings of the International Conference on Machine Learning (ICML)}, pages 448--456, 2015.

\bibitem[Kirkpatrick et~al.(2017)Kirkpatrick, Pascanu, Rabinowitz, Veness, Desjardins, Rusu, Milan, Quan, Ramalho, Grabska-Barwinska, Hassabis, Clopath, Kumaran, and Hadsell]{doi:10.1073/pnas.1611835114}
J.~Kirkpatrick, R.~Pascanu, N.~Rabinowitz, J.~Veness, G.~Desjardins, A.~A. Rusu, K.~Milan, J.~Quan, T.~Ramalho, A.~Grabska-Barwinska, D.~Hassabis, C.~Clopath, D.~Kumaran, and R.~Hadsell.
\newblock Overcoming catastrophic forgetting in neural networks.
\newblock \emph{Proceedings of the National Academy of Sciences}, 114\penalty0 (13):\penalty0 3521--3526, 2017.

\bibitem[Lee et~al.(2023)Lee, Zhong, and Wang]{Lee_2023_WACV}
K.-Y. Lee, Y.~Zhong, and Y.-X. Wang.
\newblock Do pre-trained models benefit equally in continual learning?
\newblock In \emph{Proceedings of the IEEE/CVF Winter Conference on Applications of Computer Vision (WACV)}, pages 6485--6493, 2023.

\bibitem[Li and Hoiem(2018)]{8107520}
Z.~Li and D.~Hoiem.
\newblock Learning without forgetting.
\newblock \emph{IEEE Transactions on Pattern Analysis and Machine Intelligence}, 40\penalty0 (12):\penalty0 2935--2947, 2018.

\bibitem[Liang and Li(2023)]{NEURIPS2023_249f73e0}
Y.-S. Liang and W.-J. Li.
\newblock Loss decoupling for task-agnostic continual learning.
\newblock In \emph{Proceedings of the Advances in Neural Information Processing Systems (NeurIPS)}, pages 11151--11167, 2023.

\bibitem[Liu et~al.(2021)Liu, Cai, Guo, and Chen]{Liu_Cai_Guo_Chen_2021}
B.~Liu, Y.~Cai, Y.~Guo, and X.~Chen.
\newblock {TransTailor}: Pruning the pre-trained model for improved transfer learning.
\newblock In \emph{Proceedings of the AAAI Conference on Artificial Intelligence}, pages 8627--8634, 2021.

\bibitem[Liu et~al.(2017)Liu, Li, Shen, Huang, Yan, and Zhang]{8237560}
Z.~Liu, J.~Li, Z.~Shen, G.~Huang, S.~Yan, and C.~Zhang.
\newblock Learning efficient convolutional networks through network slimming.
\newblock In \emph{Proceedings of the IEEE/CVF International Conference on Computer Vision (ICCV)}, pages 2755--2763, 2017.

\bibitem[Liu et~al.(2024)Liu, Li, Gong, and Wu]{10446458}
Z.~Liu, Y.~Li, Y.~Gong, and Y.-C. Wu.
\newblock Learning a low-rank feature representation: Achieving better trade-off between stability and plasticity in continual learning.
\newblock In \emph{Proceedings of the IEEE International Conference on Acoustics, Speech and Signal Processing (ICASSP)}, pages 5885--5889, 2024.

\bibitem[Masana et~al.(2023)Masana, Liu, Twardowski, Menta, Bagdanov, and van~de Weijer]{9915459}
M.~Masana, X.~Liu, B.~Twardowski, M.~Menta, A.~D. Bagdanov, and J.~van~de Weijer.
\newblock Class-incremental learning: Survey and performance evaluation on image classification.
\newblock \emph{IEEE Transactions on Pattern Analysis and Machine Intelligence}, 45\penalty0 (5):\penalty0 5513--5533, 2023.

\bibitem[Mensink et~al.(2013)Mensink, Verbeek, Perronnin, and Csurka]{6517188}
T.~Mensink, J.~Verbeek, F.~Perronnin, and G.~Csurka.
\newblock Distance-based image classification: Generalizing to new classes at near-zero cost.
\newblock \emph{IEEE Transactions on Pattern Analysis and Machine Intelligence}, 35\penalty0 (11):\penalty0 2624--2637, 2013.

\bibitem[Molchanov et~al.(2017)Molchanov, Tyree, Karras, Aila, and Kautz]{molchanov2017pruning}
P.~Molchanov, S.~Tyree, T.~Karras, T.~Aila, and J.~Kautz.
\newblock Pruning convolutional neural networks for resource efficient inference.
\newblock In \emph{Proceedings of the International Conference on Learning Representations (ICLR)}, 2017.

\bibitem[Ostapenko et~al.(2022)Ostapenko, Lesort, Rodriguez, Arefin, Douillard, Rish, and Charlin]{pmlr-v199-ostapenko22a}
O.~Ostapenko, T.~Lesort, P.~Rodriguez, M.~R. Arefin, A.~Douillard, I.~Rish, and L.~Charlin.
\newblock Continual learning with foundation models: An empirical study of latent replay.
\newblock In \emph{Proceedings of the Conference on Lifelong Learning Agents (CoLLAs)}, pages 60--91, 2022.

\bibitem[Panos et~al.(2023)Panos, Kobe, Reino, Aljundi, and Turner]{10378197}
A.~Panos, Y.~Kobe, D.~O. Reino, R.~Aljundi, and R.~E. Turner.
\newblock First session adaptation: A strong replay-free baseline for class-incremental learning.
\newblock In \emph{Proceedings of the IEEE/CVF International Conference on Computer Vision (ICCV)}, pages 18774--18784, 2023.

\bibitem[Petit et~al.(2023)Petit, Popescu, Schindler, Picard, and Delezoide]{petit2023fetril}
G.~Petit, A.~Popescu, H.~Schindler, D.~Picard, and B.~Delezoide.
\newblock Fetril: Feature translation for exemplar-free class-incremental learning.
\newblock In \emph{Proceedings of the IEEE/CVF Winter Conference on Applications of Computer Vision (WACV)}, pages 3911--3920, 2023.

\bibitem[Ramasesh et~al.(2022)Ramasesh, Lewkowycz, and Dyer]{ramasesh2022effect}
V.~V. Ramasesh, A.~Lewkowycz, and E.~Dyer.
\newblock Effect of scale on catastrophic forgetting in neural networks.
\newblock In \emph{Proceedings of the International Conference on Learning Representations (ICLR)}, 2022.

\bibitem[Rebuffi et~al.(2017)Rebuffi, Kolesnikov, Sperl, and Lampert]{Rebuffi_2017_CVPR}
S.-A. Rebuffi, A.~Kolesnikov, G.~Sperl, and C.~H. Lampert.
\newblock {iCaRL}: Incremental classifier and representation learning.
\newblock In \emph{Proceedings of the IEEE/CVF Conference on Computer Vision and Pattern Recognition (CVPR)}, July 2017.

\bibitem[Shon et~al.(2022)Shon, Lee, Kim, and Kim]{shon2022dlcft}
H.~Shon, J.~Lee, S.~H. Kim, and J.~Kim.
\newblock {DLCFT}: Deep linear continual fine-tuning for general incremental learning.
\newblock In \emph{Proceedings of the European Conference on Computer Vision (ECCV)}, pages 513--529, 2022.

\bibitem[Tao et~al.(2024)Tao, Yu, Yao, Huang, and Xu]{Tao2024}
Z.~Tao, L.~Yu, H.~Yao, S.~Huang, and C.~Xu.
\newblock Class incremental learning for light-weighted networks.
\newblock \emph{IEEE Transactions on Circuits and Systems for Video Technology}, 34\penalty0 (12):\penalty0 12210--12220, 2024.

\bibitem[Veniat et~al.(2021)Veniat, Denoyer, and Ranzato]{veniat:hal-03276781}
T.~Veniat, L.~Denoyer, and M.~Ranzato.
\newblock Efficient continual learning with modular networks and task-driven priors.
\newblock In \emph{Proceedings of the International Conference on Learning Representations (ICLR)}, 2021.

\bibitem[Wang et~al.(2022{\natexlab{a}})Wang, Zhou, Ye, and Zhan]{wang2022foster}
F.-Y. Wang, D.-W. Zhou, H.-J. Ye, and D.-C. Zhan.
\newblock Foster: Feature boosting and compression for class-incremental learning.
\newblock In \emph{Proceedings of the European Conference on Computer Vision (ECCV)}, pages 398--414, 2022{\natexlab{a}}.

\bibitem[Wang et~al.(2024)Wang, Zhang, Su, and Zhu]{10444954}
L.~Wang, X.~Zhang, H.~Su, and J.~Zhu.
\newblock A comprehensive survey of continual learning: Theory, method and application.
\newblock \emph{IEEE Transactions on Pattern Analysis and Machine Intelligence}, 46\penalty0 (8):\penalty0 5362--5383, 2024.

\bibitem[Wang et~al.(2021)Wang, Li, Sun, and Xu]{Wang_2021_CVPR}
S.~Wang, X.~Li, J.~Sun, and Z.~Xu.
\newblock Training networks in null space of feature covariance for continual learning.
\newblock In \emph{Proceedings of the IEEE/CVF Conference on Computer Vision and Pattern Recognition (CVPR)}, pages 184--193, 2021.

\bibitem[Wang et~al.(2022{\natexlab{b}})Wang, Zhang, Ebrahimi, Sun, Zhang, Lee, Ren, Su, Perot, Dy, et~al.]{wang2022dualprompt}
Z.~Wang, Z.~Zhang, S.~Ebrahimi, R.~Sun, H.~Zhang, C.-Y. Lee, X.~Ren, G.~Su, V.~Perot, J.~Dy, et~al.
\newblock Dualprompt: Complementary prompting for rehearsal-free continual learning.
\newblock In \emph{Proceedings of the European Conference on Computer Vision (ECCV)}, pages 631--648, 2022{\natexlab{b}}.

\bibitem[Wang et~al.(2022{\natexlab{c}})Wang, Zhang, Lee, Zhang, Sun, Ren, Su, Perot, Dy, and Pfister]{9878681}
Z.~Wang, Z.~Zhang, C.-Y. Lee, H.~Zhang, R.~Sun, X.~Ren, G.~Su, V.~Perot, J.~Dy, and T.~Pfister.
\newblock Learning to prompt for continual learning.
\newblock In \emph{Proceedings of the IEEE/CVF Conference on Computer Vision and Pattern Recognition (CVPR)}, pages 139--149, 2022{\natexlab{c}}.

\bibitem[Yan et~al.(2021)Yan, Xie, and He]{9578633}
S.~Yan, J.~Xie, and X.~He.
\newblock {DER}: Dynamically expandable representation for class incremental learning.
\newblock In \emph{Proceedings of the IEEE/CVF Conference on Computer Vision and Pattern Recognition (CVPR)}, pages 3013--3022, 2021.

\bibitem[Zhang et~al.(2020)Zhang, Sax, Zamir, Guibas, and Malik]{zhang2020side}
J.~O. Zhang, A.~Sax, A.~Zamir, L.~Guibas, and J.~Malik.
\newblock Side-tuning: a baseline for network adaptation via additive side networks.
\newblock In \emph{Proceedings of the European Conference on Computer Vision (ECCV)}, pages 698--714, 2020.

\bibitem[Zhao et~al.(2020)Zhao, Xiao, Gan, Zhang, and Xia]{9156766}
B.~Zhao, X.~Xiao, G.~Gan, B.~Zhang, and S.-T. Xia.
\newblock Maintaining discrimination and fairness in class incremental learning.
\newblock In \emph{Proceedings of the IEEE/CVF Conference on Computer Vision and Pattern Recognition (CVPR)}, pages 13205--13214, 2020.

\bibitem[Zhou et~al.(2023)Zhou, Wang, Ye, and Zhan]{zhou2023pycil}
D.-W. Zhou, F.-Y. Wang, H.-J. Ye, and D.-C. Zhan.
\newblock {PyCIL}: a python toolbox for class-incremental learning.
\newblock \emph{SCIENCE CHINA Information Sciences}, 66\penalty0 (9):\penalty0 197101, 2023.

\bibitem[Zhou et~al.(2024)Zhou, Sun, Ning, Ye, and Zhan]{ijcai2024p924}
D.-W. Zhou, H.-L. Sun, J.~Ning, H.-J. Ye, and D.-C. Zhan.
\newblock Continual learning with pre-trained models: A survey.
\newblock In \emph{Proceedings of the International Joint Conference on Artificial Intelligence (IJCAI)}, pages 8363--8371, 2024.

\bibitem[Zhu et~al.(2021)Zhu, Zhang, Wang, Yin, and Liu]{9578909}
F.~Zhu, X.-Y. Zhang, C.~Wang, F.~Yin, and C.-L. Liu.
\newblock Prototype augmentation and self-supervision for incremental learning.
\newblock In \emph{Proceedings of the IEEE/CVF Conference on Computer Vision and Pattern Recognition (CVPR)}, pages 5867--5876, 2021.

\end{thebibliography}

\clearpage
\onecolumn
\appendix
\section*{Appendix}
\addcontentsline{toc}{section}{Appendix}

\section{Baseline methods and integration of proposed KD framework}
\label{app:kd_icarl_ssil}

In this section we will briefly describe LwF, iCaRL and SS-IL that are considered as the standard CIL methods in this study.
Furthermore, we give method-specific details that how to integrate the propsed KD-based framework described in Section~\ref{sec:KD+LwF} to these methods.

For all methods, the model at task $t{=}1$ is trained using the standard classification loss:
\begin{displaymath}
    \mathcal{L}_{\text{init}} = \sum_{(x, y) \in \mathcal{D}_1}\mathcal{L}_{\mathrm{cls}}({f}_{\mathbf{\Theta_1}}(x), y) .
  \end{displaymath}
The adaptation of the proposed KD-based framework for task $t=1$ is as described in Section~\ref{sec:KD+LwF}. 
Training procedures for $t>1$ for different methods and adaptation of the proposed framework are described below.

\paragraph{LwF.}
The training loss for LwF~\cite{8107520} at task $t>1$ is given by:
\begin{equation} 
    \label{eq:replay_loss_lwf} 
    \begin{aligned} 
        \mathcal{L}_{\mathrm{LwF}}(\mathbf{\Theta}_t; \mathcal{D}_t) = \sum_{(x, y) \in \mathcal{D}_t} \Big[ &\mathcal{L}_{\mathrm{cls}}(f_{\mathbf{\Theta}_t}(x), y)+\lambda\, \text{KL} \left( \mathbf{p}_{\mathcal{C}_{t-1}}(f_{\mathbf{\Theta}_{t-1}}(x), \tau) \parallel \mathbf{p}_{\mathcal{C}_{t-1}}(f_{\mathbf{\Theta}_t}(x), \tau) \right) \Big],
    \end{aligned} 
\end{equation}
The KD-enhanced variant of LwF is described in detail in Section~\ref{sec:KD+LwF}.

\paragraph{iCaRL.}
The original training loss for iCaRL~\cite{Rebuffi_2017_CVPR} at task $t>1$ is defined as:
\begin{equation} 
    \label{eq:replay_loss_icarl} 
    \begin{aligned} 
        \mathcal{L}_{\mathrm{iCaRL}}(\mathbf{\Theta}_t; \mathcal{D}_t, \mathcal{D}_M) = \sum_{(x, y) \in \mathcal{D}_M \cup \mathcal{D}_t} \Big[ &\mathcal{L}_{\mathrm{cls}}(f_{\mathbf{\Theta}_t}(x), y)+\lambda\, \text{KL} \left( \mathbf{p}_{\mathcal{C}_{t-1}}(f_{\mathbf{\Theta}_{t-1}}(x), \tau) \parallel \mathbf{p}_{\mathcal{C}_{t-1}}(f_{\mathbf{\Theta}_t}(x), \tau) \right) \Big],
    \end{aligned} 
\end{equation}
where $\mathcal{D}_M$ is a fixed-size memory buffer storing exemplars from previous tasks. Exemplars are selected based on their $\ell_2$ distance to the class mean in the feature space.

In the proposed KD-enhanced variant of iCaRL, the teacher follows the standard iCaRL training loss. 
The student replaces the second term—the original forgetting-regularization component—with a teacher-driven distillation term. 
The overall loss is given as:
\begin{equation} 
    \label{eq:replay_loss_icarl_kd}
    \begin{aligned}
        \mathcal{L}^{\mathrm{S}}_{\mathrm{iCaRL}}(\mathbf{\Theta}_t; \mathcal{D}_t, \mathcal{D}_M) = \sum_{(x, y) \in \mathcal{D}_M \cup \mathcal{D}_t} \Big[ &\mathcal{L}_{\mathrm{cls}}(f^{\mathrm{S}}_{\mathbf{\Theta}_t}(x), y) + \lambda\, \text{KL} \left( \mathbf{p}_{\mathcal{C}_{t-1}}(f^{\mathrm{T}}_{\mathbf{\Theta}_{t-1}}(x), \tau) \parallel \mathbf{p}_{\mathcal{C}_{t-1}}(f^{\mathrm{S}}_{\mathbf{\Theta}_t}(x), \tau) \right) \Big].
    \end{aligned}
\end{equation}

\paragraph{SS-IL.}
SS-IL follows the same strategy to iCaRL. The original loss for SS-IL
for task $t>1$ is given by \cite{Ahn_2021_ICCV}:
\begin{equation} 
    \label{eq:replay_loss_ssil} 
    \begin{aligned} 
        \mathcal{L}_{\mathrm{SS\text{-}IL}}(\mathbf{\Theta}_t; \mathcal{D}_t, \mathcal{D}_M) =\!\!\! &\sum_{(x, y) \in \mathcal{D}_t} \mathcal{L}_{\mathrm{cls}}\left(\mathbf{p}_{\mathcal{C}_t}(f_{\mathbf{\Theta}_t}(x)), y \right)+\!\!\!\sum_{(x, y) \in \mathcal{D}_M} \mathcal{L}_{\mathrm{cls}}\left(\mathbf{p}_{\mathcal{C}_{t-1}}(f_{\mathbf{\Theta}_t}(x)), y \right) \\
        +\!\!\! &\sum_{(x, y) \in \mathcal{D}_t \cup \mathcal{D}_M} \lambda\, \text{KL} \left( \mathbf{p}_{\mathcal{C}_{t-1}}(f_{\mathbf{\Theta}_{t-1}}(x), \tau) \parallel \mathbf{p}_{\mathcal{C}_{t-1}}(f_{\mathbf{\Theta}_t}(x), \tau) \right),
    \end{aligned} 
\end{equation}
where $\mathcal{C}_t$ denotes the set of current task classes, and $\mathcal{C}_{t-1}$ the set of classes from all previous tasks.

In the KD-enhanced variant of SS-IL, the teacher also follows the original SS-IL training loss. The student replaces the third term—originally a self-distillation component aimed at mitigating forgetting—with a teacher-driven distillation term. The overall loss is given as:
\begin{equation} 
    \label{eq:replay_loss_ssil_kd}
    \begin{aligned} 
        \mathcal{L}^{\mathrm{S}}_{\mathrm{SS\text{-}IL}}(\mathbf{\Theta}_t; \mathcal{D}_t, \mathcal{D}_M) =\!\!\! &\sum_{(x, y) \in \mathcal{D}_t} \mathcal{L}_{\mathrm{cls}}\left(\mathbf{p}_{\mathcal{C}_t}(f_{\mathbf{\Theta}_t}(x)), y \right) +\!\!\! \sum_{(x, y) \in \mathcal{D}_M} \mathcal{L}_{\mathrm{cls}}\left(\mathbf{p}_{\mathcal{C}_{t-1}}(f_{\mathbf{\Theta}_t}(x)), y \right) \\
        +\!\!\! &\sum_{(x, y) \in \mathcal{D}_t \cup \mathcal{D}_M} \lambda\, \text{KL} \left( \mathbf{p}_{\mathcal{C}_{t-1}}(f^{\mathrm{T}}_{\mathbf{\Theta}_{t-1}}(x), \tau) \parallel \mathbf{p}_{\mathcal{C}_{t-1}}(f^{\mathrm{S}}_{\mathbf{\Theta}_t}(x), \tau) \right).
    \end{aligned} 
\end{equation}

\section{ACC vs. task number on OoD datasets for the proposed KD framework}
\label{app:ood_acc}

\begin{figure}[ht]
    \centering
    \includegraphics[width=0.6\textwidth]{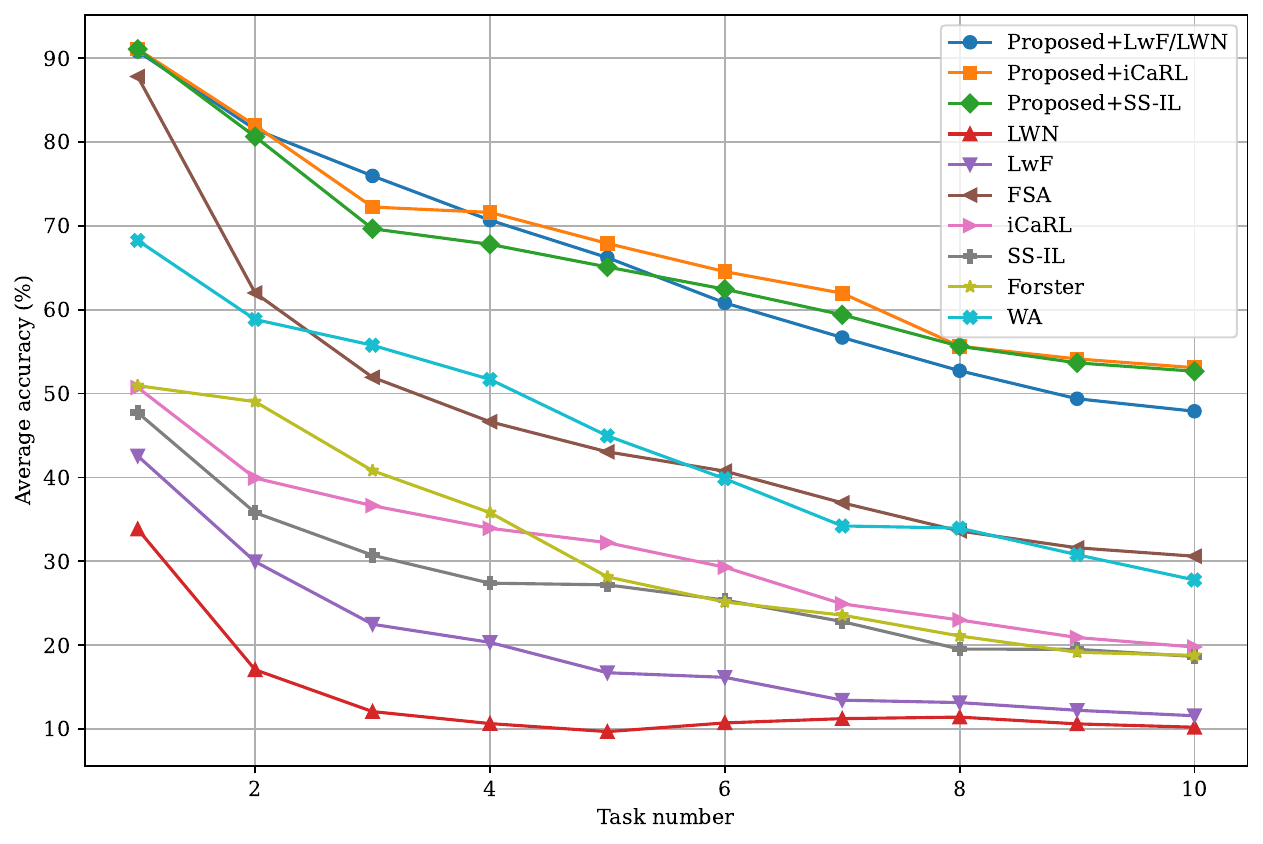}
    \caption{ACC vs. task number on FGVC Aircrafts for the proposed KD framework.} 
    \label{task_accuracy_aircrafts-10}
\end{figure}

\begin{figure}[ht]
    \centering
    \includegraphics[width=0.6\textwidth]{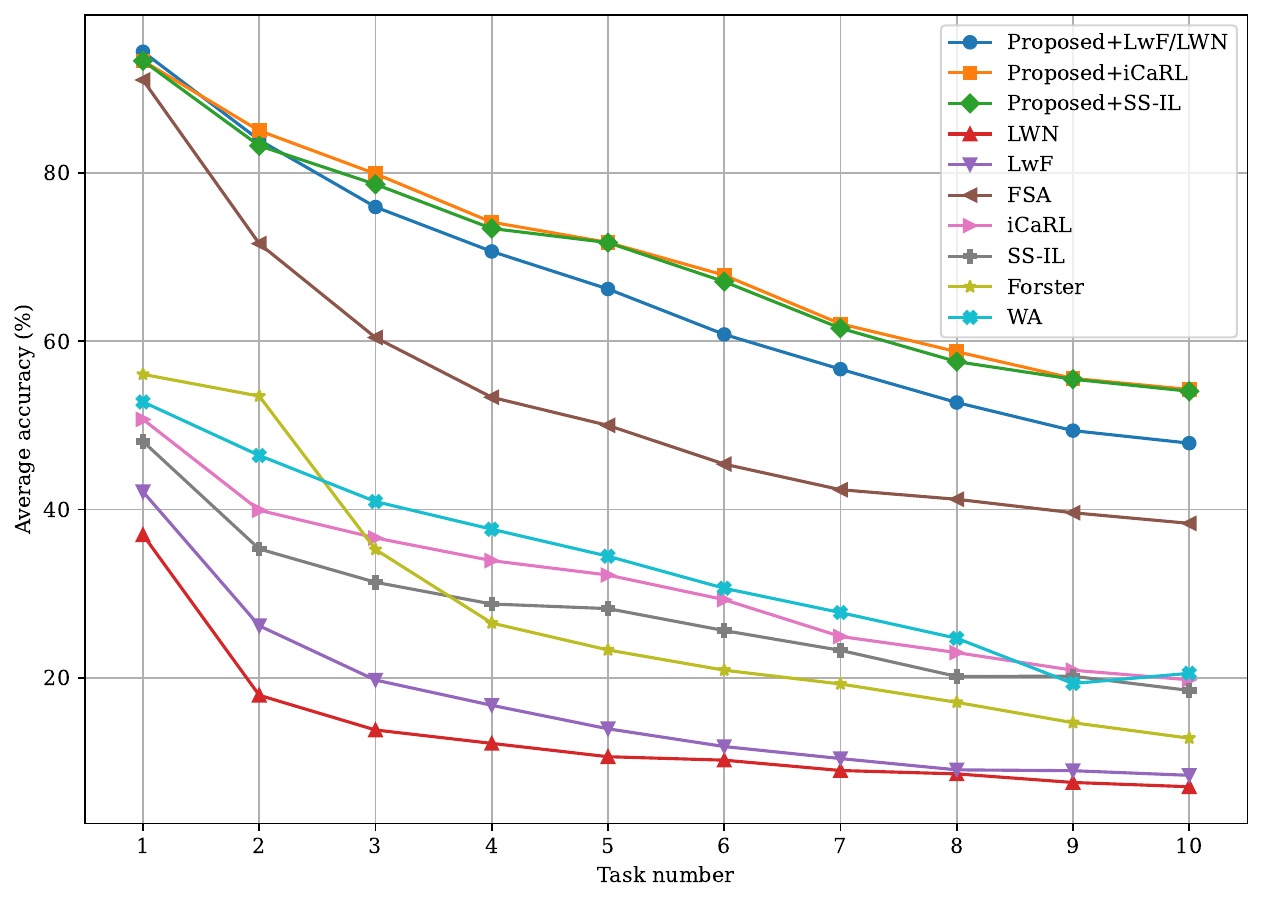}
    \caption{ACC vs. task number on Cars for the proposed KD framework.} 
    \label{task_accuracy_cars-10}
\end{figure}

Figures~\ref{task_accuracy_aircrafts-10} and \ref{task_accuracy_cars-10} illustrate the progression of ACC across incremental tasks on the FGVC Aircraft and Stanford Cars datasets.
These plots complement the OoD performance analysis in Table~\ref{tab:combined_results}, as discussed in Section~\ref{sec:kd-based experiment}.
As noted there, limited data makes it challenging for LWN to train its large teacher model from scratch effectively, which is reflected in both its low ACC and small absolute BWT.
This trend is further corroborated by the figures: the substantial accuracy gap between LWN and Proposed+LWN after the first task highlights LWN’s poor task-wise accuracy resulting from data insufficiency.

The figures also help clarify the observed BWT differences between baselines and proposed methods, as explained in Section~\ref{sec:kd-based experiment}.
Specifically, baseline methods (e.g., LWN) start with lower accuracy on the first task, which naturally leads to smaller subsequent drops and thus artificially inflated BWT scores.
In contrast, KD-based methods (e.g., Proposed + LWN), benefiting from the teacher’s generalization capability, achieve much higher accuracy after the first task.
Although this may result in larger drops—and therefore lower BWT—it actually reflects stronger overall performance rather than increased forgetting.

\end{document}